# mcp-proto-okn: Natural-language access to open scientific knowledge graphs through the Model Context Protocol

Peter W. Rose[1], Benjamin M. Good[2], Amanda M. Saravia-Butler[3], Charlotte A. Nelson[4], James P. Balhoff[5], Yaphet Kebede[5], Patricia L. Whetzel[2], Christopher Bizon[5], Andrew I. Su[2], Sergio E. Baranzini[6]

[1]San Diego Supercomputer Center, University of California San Diego, La Jolla, CA 92093, USA
[2]The Scripps Research Institute, La Jolla, CA 92037, USA
[3]Amentum, Space Biosciences Division, NASA Ames Research Center, Moffett Field, CA 94035, USA
[4]Mate Bioservices, Redwood City, CA 94065, USA
[5]Renaissance Computing Institute, University of North Carolina at Chapel Hill, Chapel Hill, NC 27517, USA
[6]Weill Institute for Neurosciences, Department of Neurology, University of California San Francisco, San Francisco, CA 94158, USA

## ABSTRACT

### Summary

MCP Server Proto-OKN (`mcp-proto-okn`) is a Python-based Model Context Protocol server that enables AI assistants to discover, inspect, query and integrate scientific knowledge graphs through natural language. The server provides graph routing, schema inspection, SPARQL execution, ontology expansion, multi-graph querying, and transcript generation, lowering the barrier to cross-domain knowledge graph analysis for biomedical and scientific users.

### Availability and Implementation

`mcp-proto-okn` is implemented in Python using the FastMCP framework and is available at https://github.com/sbl-sdsc/mcp-proto-okn. Documentation, client configuration instructions, and example analysis transcripts are provided in the GitHub repository.

Contact: pwrose.ucsd@gmail.com, sergio.baranzini@ucsf.edu

## INTRODUCTION

The rapid proliferation of biomedical and scientific data has driven the development of knowledge graphs (KGs) as a powerful paradigm for representing, integrating, and querying heterogeneous information across biological, clinical, and environmental domains. By encoding semantically linked entities and relationships, KGs enable reasoning across disciplinary boundaries, connecting genes to diseases, chemicals to toxicological outcomes, and genomic perturbations to clinical phenotypes. Recognizing this potential, the U.S. National Science Foundation (NSF) invested $26.7 million in the "Building the Prototype Open Knowledge Network (Proto-OKN)" program (https://www.proto-okn.net/), funding multidisciplinary projects to construct an interconnected network of open, publicly accessible knowledge graphs spanning domains from biomedicine and environmental science to criminal justice and supply chain logistics. These graphs are hosted and interconnected through the OKN Fabric (https://registry.okn.us/), which provides the technical infrastructure to link many individual graphs into a coherent, queryable network.

Despite this infrastructure, programmatic access to the full Proto-OKN ecosystem remains a significant challenge for end users. The 30+ constituent knowledge graphs span different ontological domains, share few entities, and expose heterogeneous SPARQL endpoints requiring domain-specific expertise. This fragmentation limits integrative analyses across graphs, particularly for researchers without SPARQL expertise.

The emergence of the Model Context Protocol (MCP) (https://modelcontextprotocol.io/), a standard for enabling large language model (LLM)-based AI assistants to interact with external tools and data services, now offers a compelling solution to this access barrier. By exposing structured query capabilities as MCP tools, knowledge graphs become queryable through natural language, enabling AI assistants to construct and execute SPARQL queries on behalf of users, interpret results, and chain queries across multiple endpoints in a single conversational session.

Recent work has explored using LLMs to translate natural-language questions into SPARQL queries for life-science knowledge graphs. For example, Emonet *et al.* 2025 developed a retrieval-augmented generation (RAG) framework for generating federated SPARQL queries over the SIB Expasy bioinformatics knowledge graphs, using schema metadata, curated question–query pairs, and a validation step to improve accuracy and reduce hallucinations. TogoMCP introduced an MCP-based framework in which LLMs orchestrate tool calls to inspect RDF schemas, resolve entities through external REST APIs, and generate SPARQL queries against the DBCLS RDF Portal, which aggregates over 70 life-science databases under shared schema conventions (Kinjo *et al.* 2026). Both efforts target curated ecosystems. The Proto-OKN setting differs: over 30 graphs span biomedical, environmental, and other domains, share few entities, and follow no common schema convention — requiring routing between graphs, hierarchical reasoning over reference ontologies, and provenance tracking across multi-graph queries. Here, we present `mcp-proto-okn`, a Python-based MCP server that addresses these requirements for the Proto-OKN ecosystem on the OKN Fabric. Beyond text-to-SPARQL generation, it supports graph routing, ontology expansion with UberGraph (https://github.com/INCATools/ubergraph/), coordinated multi-graph querying, and creating chat transcripts. We provide illustrative use cases spanning spaceflight gene expression analysis, cross-domain data exploration, and dataset retrieval with ontology expansion. An example workflow is depicted in Figure 1.

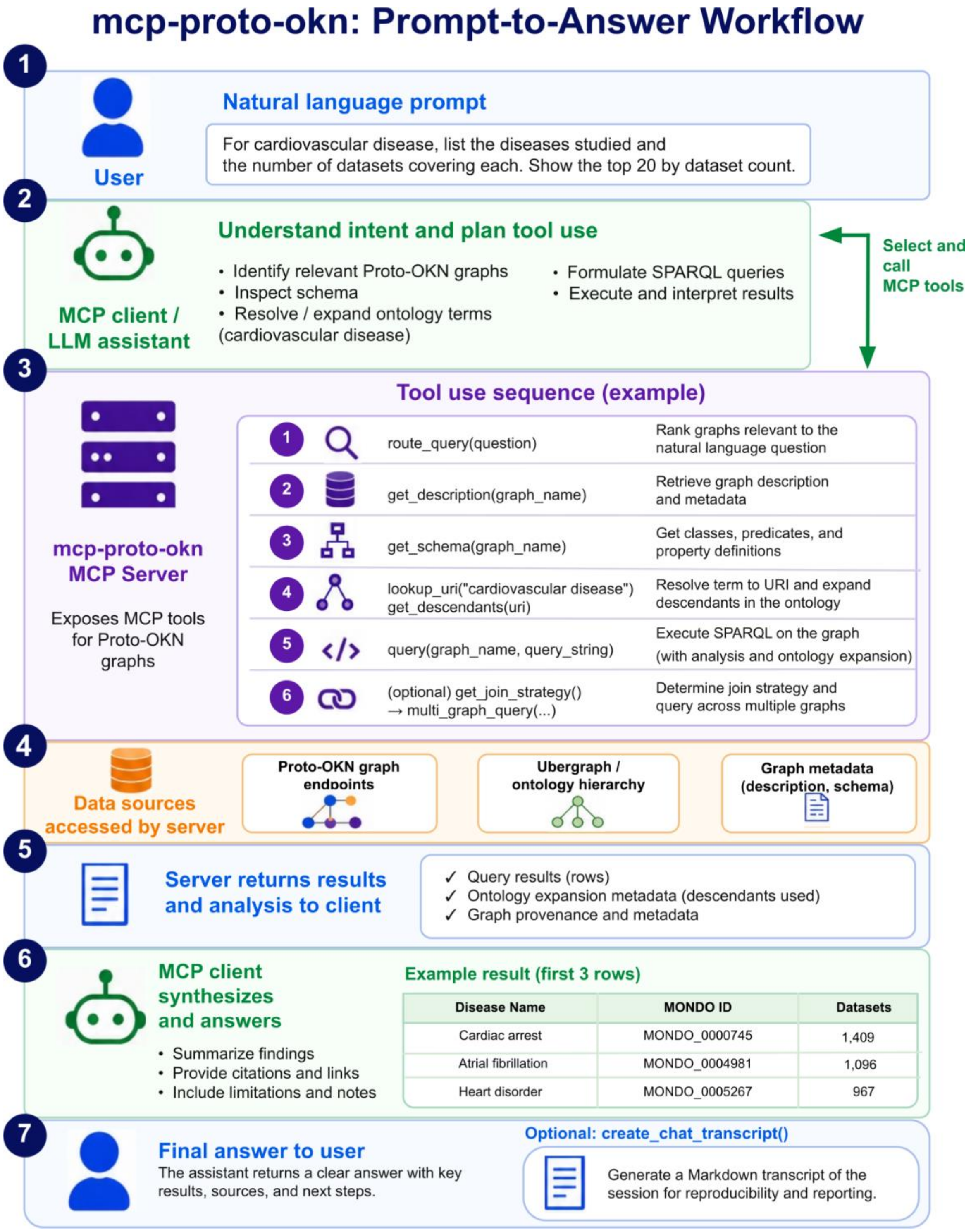

**Figure 1. MCP-enabled prompt-to-answer workflow for Proto-OKN graph queries.**
A user submits a natural-language question, which the MCP client/LLM assistant translates into graph selection, schema inspection, ontology expansion, and SPARQL query execution through the unified `mcp-proto-okn` server. The server returns results and metadata, which the assistant synthesizes into a final response with key findings and sources.

## METHODS

The Proto-OKN knowledge graphs are hosted on the OKN Fabric and are exposed at the SPARQL endpoint https://frink.apps.renci.org/federation/sparql. Each graph occupies a distinct named graph URI (https://purl.org/okn/frink/kg/<name>) and is cataloged with metadata and documentation links. The complete Proto-OKN registry is available at https://registry.okn.us/registry/.

The biomedically relevant knowledge graphs include SPOKE-OKN (genes, diseases, compounds, social determinants of health) (Morris *et al.* 2023), SPOKE-GeneLab (NASA GeneLab spaceflight omics data) (Gebre *et al.* 2025), Gene Expression Atlas OKN (Madrigal *et al*. 2026), the BioBricks toxicology graphs (Gao *et al.* 2024), NDE (NIAID Data Ecosystem) (Tsueng *et al.* 2026), and ProKN (Protein Knowledge Network, https://research.bioinformatics.udel.edu/ProKN). Additional knowledge graphs are being added to the OKN Fabric and will become available through `mcp-proto-okn` as they are integrated.

`mcp-proto-okn` is implemented in Python using the FastMCP framework, with decorator-based tool registration. All Proto-OKN graphs are registered simultaneously, and users can reference graphs by name (e.g., @spoke-genelab) in natural-language prompts. Users configure the server by adding an MCP URL to an MCP-compatible client, such as Claude Desktop, ChatGPT, or another compatible assistant (https://github.com/sbl-sdsc/mcp-proto-okn#connecting-your-client).

The server exposes tools organized into five functional categories. Knowledge graph discovery tools identify relevant data sources before query construction. `list_graphs` summarizes available graphs, including metadata such as domains, entity types, and identifier namespaces, while `route_query` suggests candidate graphs for a natural-language question and `get_description` provides graph-specific details and examples.

Schema and query-construction tools support valid SPARQL generation. `get_schema` returns graph classes, predicates, and properties for query construction. `get_query_template` provides patterns for relationships that require specialized SPARQL structures, such as queries over edge properties using RDF reification.

Query tools execute SPARQL against one or more graphs. `query` executes graph-scoped SPARQL and returns structured tabular results along with query-analysis metadata and warnings about common issues. `multi_graph_query` supports coordinated execution across multiple graphs and annotates results with their source graph.

Ontology and federation tools support hierarchical and cross-graph analysis. The `query` tool can automatically expand ontology identifiers, such as MONDO, UBERON, HP, GO, CL, and ChEBI terms, by retrieving descendants from the `UberGraph KG` and injecting them into the SPARQL query. `lookup_uri` and `get_descendants` support ontology exploration, while `get_join_strategy` recommends how to combine graphs using shared identifier namespaces or bridge graphs.

Finally, documentation and visualization tools support reproducibility and reporting. `visualize_schema` assists with schema visualization, and `create_chat_transcript` generates a structured Markdown record of an analysis session.

The AI assistant handles identifier conversion, result merging, and cross-graph reasoning, making federation robust to endpoint heterogeneity and schema differences. The use cases described below additionally leverage the PubMed MCP Server, illustrating that MCP Server Proto-OKN integrates naturally with the broader ecosystem of scientific MCP services.

## RESULTS

*Case study 1: Spaceflight-associated changes in thymus gene expression*

To illustrate the integrative analytical potential of `mcp-proto-okn`, we demonstrate a multi-step spaceflight biology workflow conducted entirely through natural language queries. The analysis drew on two complementary knowledge graphs: spoke-genelab, which encodes NASA GeneLab experimental data including differential expression results from spaceflight studies, and spoke-okn, which provides broad biomedical context including gene–disease associations.

For this example, we analyzed the NASA GeneLab study OSD-244, a Rodent Research-6 spaceflight study from the SpaceX-13 mission that profiled Mus musculus thymus by RNA-seq (Chat transcript: https://github.com/sbl-sdsc/mcp-proto-okn/blob/main/docs/examples/spoke-genelab-OSD-244_verbatim.md). The assistant first queried the spoke-genelab KG to recover study metadata and assay design, identifying matched Space Flight versus Ground Control comparisons at ~30 and ~60 days. Differential-expression analysis revealed thousands of significant genes, with stronger ~60-day signatures related to coagulation, extracellular matrix remodeling, metabolism, and muscle-associated pathways. Cross-endpoint comparison identified 322 genes significantly altered at both time points, with strong concordance between conditions (Pearson correlation coefficient = 0.80), indicating a sustained spaceflight-associated transcriptional response.

The assistant then mapped mouse genes to human orthologs and switched to the spoke-okn KG to retrieve disease associations, identifying associations with cardiovascular, metabolic, inflammatory, autoimmune, neurological, coagulation, and liver diseases. The workflow also generated assay-design diagrams, concordance plots, disease-association summaries, and literature-supported interpretations. This case study demonstrates how mcp-proto-okn enables progression from study discovery to differential-expression analysis, cross-species translation, disease contextualization, visualization, and interpretation through natural-language interaction alone.

*Case study 2: Automatic ontology expansion for cross-ontology dataset discovery*

A second use case illustrates a recurring challenge in querying biomedical knowledge graphs: terms in users' questions are typically high-level concepts (e.g., "cardiovascular disease"), whereas datasets and records in the underlying graphs are annotated with specific subtype identifiers drawn from large reference ontologies. A query filtering on a single root URI misses most relevant records.

We demonstrate this with the NIAID Data Ecosystem (NDE), a discovery catalog of biomedical datasets in which each dataset is tagged with one or more MONDO disease URIs (Chat transcript: https://github.com/sbl-sdsc/mcp-proto-okn/blob/main/docs/examples/cardiovascular-disease-ontology-expansion.md). In response to the natural-language prompt *"@NDE For cardiovascular disease, list the diseases studied and the number of datasets covering each; show the top 20 by dataset count,"* the assistant first mapped the term "cardiovascular disease" to a canonical URI (MONDO:0004995, corresponding to "cardiovascular disorder") (Figure 1).

A naïve approach would then issue a single-URI SPARQL query against the NDE graph. This approach yielded 447 datasets annotated directly with the root-level "cardiovascular disorder" term. The low number underscores that the bulk of relevant records are not annotated with this high-level disease term.

In contrast, our ontology-aware assistant first automatically fetched descendants of MONDO:0004995 from UberGraph and rewrote the SPARQL FILTER clause to accept any of them — expanding one input URI into 1,592 descendant URIs across the cardiovascular subtree. The expanded URI set was queried in 80 batches to respect endpoint limits, and the expansion metadata was returned to the assistant.

The resulting tabulation covered 284 distinct cardiovascular conditions that collectively retrieved from over 10,000 individual datasets. The most heavily represented disease subtypes include cardiac arrest (1,409 datasets), atrial fibrillation (1,096), heart disorder (967), atherosclerosis (859), hypertensive disorder (793), heart failure (677), myocardial infarction (525), and stroke (520). Notably, the descendant set also surfaced rare Mendelian cardiovascular syndromes (e.g., Holt-Oram, Loeys-Dietz, Brugada) that a flat keyword search would likely have missed.

The same mechanism applies automatically to identifiers from MONDO, UBERON, HP, GO, CL, and ChEBI on any graph that uses these namespaces, with configurable expansion bounds. This case study shows how ontology expansion bridges user-level concepts and curator-level annotations while preserving recall across the full ontological subtree.

## CONCLUSIONS

`mcp-proto-okn` demonstrates how the Model Context Protocol can connect fragmented scientific knowledge graphs hosted on the OKN Fabric and make them accessible to researchers without query-language expertise. Built on the OKN Fabric for graph discovery and SPARQL querying, `mcp-proto-okn` enables AI assistants to route natural-language questions to relevant Proto-OKN knowledge graphs, inspect schemas, execute queries, and synthesize results. The use cases presented here showcase its utility for cross-graph queries, ontology-expanded search, and integration with complementary MCP services such as PubMed. Together, these capabilities lower the barrier for integrative discovery across biomedicine and beyond.

Current limitations include dependence on the availability and schema quality of upstream OKN Fabric endpoints, and variation in identifier coverage across graphs. `mcp-proto-okn` mitigates these issues through schema-inspection tools, query templates, warning metadata, and provenance-preserving outputs.

## FUNDING

This work was supported by the National Science Foundation Award No. 2333819: "Proto-OKN Theme 1: Connecting Biomedical information on Earth and in Space via the SPOKE knowledge graph" and Award No. 2535091: "Proto-OKN Theme 2: OKN-Fabric".

## REFERENCES

Emonet V, Bolleman J, Duvaud S, et al. LLM-based SPARQL Query Generation from Natural Language over Federated Knowledge Graphs. 2025 *arXiv*:2410.06062. doi: 10.48550/arXiv.2410.06062

Gao Y, Mughal Z, Jaramillo-Villegas JA, et al. BioBricks.ai: A Versioned Data Registry for Life Sciences Data Assets. 2024 *arXiv*:2408.17320. doi: 10.48550/arXiv.2408.17320

Gebre SG, Scott RT, Saravia-Butler AM, et al. NASA open science data repository: open science for life in space. *Nucleic Acids Research* 2025;53(D1):D1697–D1710. doi: 10.1093/nar/gkae1116

Kinjo AR, Yamamoto Y, Bustamante-Larriet S, et al. TogoMCP: Natural Language Querying of Life-Science Knowledge Graphs via Schema-Guided LLMs and the Model Context Protocol. *bioRxiv*:2026.03.19.713030. doi: 10.64898/2026.03.19.713030

Madrigal P, Thanki AS, Fexova S, et al. Expression Atlas in 2026: enabling FAIR and open expression data through community collaboration and integration. *Nucleic Acids Research.* 2026;54(D1):D147–D157. doi:10.1093/nar/gkaf1238

Morris JH, Soman K, Akbas RE, et al. The scalable precision medicine open knowledge engine (SPOKE): a massive knowledge graph of biomedical information. *Bioinformatics* 2023;39(2):btad080. doi: 10.1093/bioinformatics/btad080

Tsueng G, Bullen E, Czech C, et al. The NIAID Discovery Portal: a unified search engine for infectious and immune-mediated disease datasets. *mSystems*. 2026;11(2):e0127025. doi:10.1128/msystems.01270-25